# Research Highlights

- Day-ahead peak load forecasting is implemented with multiple input variables.
- A novel hybrid method has been proposed based on MEMD-PSO-SVR.
- MEMD approach in multivariate data decomposition prevents information loss.
- PSO is utilized to optimize the SVR parameters.
- Power planer can use this new hybrid model for day-ahead peak load forecasting.

# Multivariate Empirical Mode Decomposition based Hybrid Model for Day-ahead Peak Load Forecasting

Yanmei Huang[1], Najmul Hasan[2], Changrui Deng[1], Yukun Bao [2]*

[1] *Center for Big Data Analytics, Jiangxi University of Engineering, Xinyu 338029, P.R. China.*

[2] *Center for Modern Information Management, School of Management, Huazhong University of Science and Technology, Wuhan, 430074, P.R. China.*

**Abstract**

Accurate day-ahead peak load forecasting is crucial not only for power dispatching but also has a great interest to investors and energy policy maker as well as government. Literature reveals that 1% error drop of forecast can reduce 10 million pounds operational cost. Thus, this study proposed a novel hybrid predictive model built upon multivariate empirical mode decomposition (MEMD) and support vector regression (SVR) with parameters optimized by particle swarm optimization (PSO), which is able to capture precise electricity peak load. The novelty of this study mainly comes from the application of MEMD, which enables the multivariate data decomposition to effectively extract inherent information among relevant variables at different time frequency during the deterioration of multivariate over time. Two real-world load data sets from the New South Wales (NSW) and the Victoria (VIC) in Australia have been considered to verify the superiority of the proposed MEMD-PSO-SVR hybrid model. The quantitative and comprehensive assessments are performed, and the results indicate that the proposed MEMD-PSO-SVR method is a promising alternative for day-ahead electricity peak load forecasting.

* Corresponding Author: Tel: +86-27-62559765; fax: +86-27-87556437.
E-mail: yukunbao@hust.edu.cn or y.bao@ieee.org

## 1. Introduction

Day-ahead peak load forecasting is a crucial requirement for load management, and it aids in the installation and operation of the power system in the short term. The utility companies must have forewarning of expected peak loads that can monitor, control the smart grid and demand response. As electrical energy cannot be stored in large quantities, utility providers' primary concern is optimizing the amount of electricity they can generate and minimizing production costs [1, 2]. In this rare instance, accurate peak load forecasting can help utility providers optimize unit production scheduling and the amount of energy supplied to the grid [3]. In some cases, a realistic forecasting model will make it easier for energy companies to analyze their economic viability within a competitive market setting [4]. It is estimated that the utility's operating costs were reduced by over 10 million pounds due to a 1% error drop in the load forecasting model [1, 5]. However, peak load forecasting is complicated and sensitive due to its complex non-linear characteristics. Non-linearity and various stochastic characteristics of electricity demand lead to the imbalance of supply and demand during the peak load period, which increases the operation cost of the power system. Therefore, predicting peak load is economically and environmentally meaningless if it is not sufficiently precise[6].Thus, a robust peak load forecast model is a prerequisite to energy market players to secure the power system's technological and economic operation [7].

In recent decades, electricity load forecasting has been classified into four forms according to the forecasting time windows: long-term[8, 9], medium-term[10, 11], short-term[2, 12], and ultra-short-term load forecasting[13]. Short-term load forecasting (STLF) is usually used to perform one-day ahead to several-weeks ahead forecasting of electricity load [14]. Compared to medium and long-term forecasting, short-term forecasting is in greater need of accuracy to make a successful decision in power grid dispatching [2]. As a result, the predictive accuracy of STLF is becoming increasingly significant [15]. Accurate STLF has great importance in power system operations such as power scheduling, power planning, and economic operations to make the system more secure and stable [16]. Thus, the main focus of this study is the STLF for day-ahead peak load forecasting for unit-commitment in short-ahead power grid optimization and accurate electricity distribution systems.

Numerous scholars have proposed different peak load forecasting models to strengthen forecasting accuracy that are mainly categorized into two sub-streams: univariate and multivariate. While the univariate technique considers only the time-series approach, the multivariate modeling process considers a range of variables, i.e., population growth, gross domestic production (GDP), and meteorological data for peak load forecasting [9]. The behavior for a load time series flow is usually complicated. The non-linear and non-stationary form of load time series leads to the complexities of peak load forecasting [17, 18]. Therefore, developing a reliable forecasting model to predict short-term peak loads with high precision is very useful because of the load signal's complexity. A variety of models were developed in an attempt to predict the amount of electricity peak load. For example, traditional statistics methods, such as linear regression[19, 20], gray forecasting [21], fuzzy logic[22, 23], exponential smoothing (ES)[24, 25], and autoregressive integrated moving average[26], are useful in forecasting linear trends. However, these approaches can't predict the non-linear signals and time series accurately due to the defect of capturing the significant fluctuation of electricity demand and they can only discuss the physical properties of the substances [27]. Similarly, the traditional prediction models produce poor results on days with abnormal or special events, such as holidays and weekends [16]. However, advanced computational models do better at predicting non-linear time series. Thus, scientists have switched from traditional mathematical analysis approaches to artificial-intelligence based computing technology to perform tasks more accurately using numerous techniques such as artificial neural networks (ANN) [2, 18, 20, 27, 28]and support vector machines (SVMs)[29-32]. Considering poor prediction result of univariate time series and inherent limitations of single model, machine learning and hybrid approaches are regarded as powerful techniques by researchers to deal with the non-linear and non-stationary characteristics of the peak load. Different short-term peak load forecasting model shave presented in Table 1.

Table 1
Different short-term peak load forecasting models established.

| References | Forecasting Strategy | Time Scale | Description |
| --- | --- | --- | --- |
| Granderson, Sharma [33] | Electricity peak load was evaluated using meter-based modeling for commercial buildings while demand response (DR) application has used | Hourly | This paper examined whether a regression model that used for predicting the hourly energy consumption is also accurate in predicting short-term peak loads. Thus, an examination of the eight different algorithms was conducted, and the findings are included in the report. |
| Li, Ma [34] | Proposed a hybrid technique using cluster analysis, Cubist regression models and Particle Swarm Optimization for day-ahead peak electricity demand of a building portfolio | Daily | First, the daily electricity usage was grouped by the use of cluster analysis with a hybrid dissimilarity approach. To enhance the prediction accuracy, the Cubist-based prediction model was trained by the clustering output. To further increase prediction performance, a Particle Swarm Optimization technique was employed to identify the optimal parameters for the cluster analysis. |
| Sakurai, Fukuyama [35] | An ANN was applied using differential evolutionary PSO (DEEPSO) considering outliers | Daily | In this paper, a novelty load forecasting algorithm of DEEPSO is proposed to predict daily peak loads by considering outliers. Even if outliers exist in training results, DEEPSO based ANN will be able to forecast better than traditional SGD and PSO based ANN's. |
| Moral-Carcedo and Pérez-García [36] | The relation between the economic activity and temperature can be evaluated using multiple linear regression(MLR) for peak load forecasting. | Hourly | A novel forecasting technique is being proposed. This one improved upon the multivariate methods and can integrate long and short-term forecast knowledge. This technique has many advantages, such as controlling for longer-term trends, such as changes in GDP growth, demographics, and technological progress. |
| Moazzami, Khodabakhshian [6] | Employed a hybrid approach incorporating wavelet decomposition and genetic algorithm training optimization of artificial neural network with weather data. | Daily | Day-ahead peak consumption is estimated by evaluating the low and high-frequency dimensions of an ANN. The findings suggest the efficacy and superiority of the proposed technique compared to other peak load forecasting approaches. |
| Haq, Lyu [37] | This proposal combines SVM and ANN with faster K-medoids clustering as an intuitively-based approach for forecasting the peak demand. | Short-term (30 min) | The K-medoids clustering approach was used to acquire three separate clusters of the experimental dataset. The clusters were used in the later process for the classification of the dataset. After a multi-step procedure, the best product features are selected and then grouped by the classification system. |

Due to the advantage of data preprocessing procedures in extracting the inherent characteristics of data series over time, the hybrid modeling frameworks employing time-frequency analysis were supposed to be promising by researchers. Through time-frequency analysis, the relationship of the two most important physical quantities (frequency domain and time domain) can clearly been presented for mining the data feature. The hybrid modeling approach enhances on current decomposition-ensemble modeling paradigms depending on the data properties of the original series, which is an improvement over previous approaches. Therefore, to improve the forecasting accuracy by decomposing the load series for capturing the complicated features at different frequencies, Moazzami, Khodabakhshian [6] employed a wavelet decomposition and genetic algorithm training optimization of artificial neural network with weather data for day-ahead peak load by evaluating the low and high-frequency dimensions of an ANN. However, the criterion of wavelet selecting basis function in advance is also quite complex. To overcome the defect of wavelet, the empirical mode decomposition (EMD) is drawing a growing deal of attention in recent years [38]. For instance, Al-Musaylh, Deo [31] used improved empirical mode decomposition with adaptive noise (ICEEMDAN) combined with support vector regression (SVR) optimized by two-phase particle swarm (PSO) algorithm for electricity demand forecasting. Recently, research on EMD has developed theoretical treatments for bivariate [39], trivariate [40], and multivariate patterns [41], respectively. Multivariate EMD (MEMD) is a multivariate and multiscale decomposition approach derived from EMD [38, 41]. EMD and MEMD deal with non-linear and non-stationary structures, which can decompose original data into intrinsic mode functions (IMFs). However, as mentioned earlier, various factors influence electricity load, such as meteorological variables, weekdays, holidays, and customer social behavior [42, 43]. The power system's complexity suggests that it is insufficient to consider the univariate historical load for peak load forecasting. The advantages of MEMD are taken into account in this study to deal with non-linear and non-stationary load series and their influencing factors. Consequently, to obtain more accurate predictions, multivariate empirical mode decomposition (MEMD) was proposed by Rehman and Mandic [41]to decompose the relational multi-variables of electricity load simultaneously, resulting in improved performance using EMD. However, it is challenging to take account of all the factors affecting the load. Based on the extensive literature analysis [43-46], the temperature is an extremely critical factor among meteorological variables. The temperature data have been added with historical loads as input variables in this study to reduce hybrid models' computational cost. The MEMD algorithm decomposed the multi-dimensional input variables to extract similar-frequency multi-dimensional intrinsic mode functions (imfs) and a multi-dimensional residue. The one day ahead peak load was the output variable or the forecast target.

Due to the strong theoretical foundation and generalization ability of support vector regression (SVR) [43, 45, 47-51], SVR was employed as modeler in this study. Another popular modeler, artificial neural network (ANN) is employed for the purpose of comparison. To further improve the forecast accuracy, particle swarm optimization (PSO) is used to optimize SVR and ANN's hyperparameters. However, it is worth noting that the focus of this research is on the validity of the proposed hybrid method rather than comparing the variant implementations with different optimization methods, such as genetic algorithm GA [6], MABC [52], FOA [53], and CLPSO [43].The proposed MEMD-based hybrid model is defined as MEMD-PSO-SVR in this study. The real-world electricity load data sets from the New South Wales (NSW) and the Victoria (VIC) in Australia have been considered to justify the proposed model's performance against other comparison models. Results in this study have demonstrated that the proposed MEMD-PSO-SVR hybrid model excels in day-ahead peak load forecasting accuracy.

The main contributions of this study are outlined as follows.

1. A novel hybrid modeling framework of day-ahead peak load forecasting has been modeled considering multi-dimensional input variables for effectively making stochastic scheduling to reduce the loss of underestimated or overestimated in power system. Although several researches have been conducted to examine the forecasting reliability and effectiveness, no studies, evidence from table 1, related to peak load forecasting using MEMD for peak load data decomposition along with SVR have been reported in the literature.

2. Unlike other decomposition techniques, the MEMD approach decomposes simultaneously history loads and meteorological variable, and adaptively handles the non-linearity and non-stationarity of electricity peak load, and effectively extracts various features at different levels of time frequencies associated with the prediction of next day's peak load more accurately. By using the MEMD decomposition approach, the well-established EEMD modeling approach can be extended to predict peak load, which not only takes advantage of the simplicity of modeling process but also avoids having to incur higher computing costs.

3. A conventional SVR model is incapable of tuning the optimal parameters that hinders the model's predictive performance. To address this shortcoming, the PSO algorithm employed to enhance the prediction capability of the SVR model by means of optimizing the model's hyperparameters. However, the proposed MEMD-PSO-SVR novel hybrid model notably produces higher significant insight, as verified by the DM test, which serves as a foundation for future energy research.

4. Another contribution is to provide first-ever strong empirical evidence by using data from two real-world electricity peak load (NSW and Victoria) cases in the electricity peak load prediction literature that a short-term robust SVR (optimized by PSO) modeling technique is integrated with the data preprocessing of time-frequency analysis method for multivariate, which extended the literature of the electricity load forecasting.

5. Power planer can use this new efficient hybrid method for day-ahead peak

load forecasting that focuses on improving the quality of input data of forecasting model instead of using complex forecasting models. Therefore, we hope this study would contribute to the analytics pool for energy analysis community.

The rest sections of this paper are organized as follows. Section 2 introduces the basic methodologies of MEMD, SVR, PSO, as well as the proposed hybrid modeling framework. Section 3 presents the experimental research in detail. The experimental results and analysis are reported in Section 4. Section 5 presents the summaries and discussion of the proposed modeling framework. Finally, the conclusions of this study are described in section 6.

## 2. Methodologies

### 2.1 Multivariate empirical mode decomposition (MEMD) algorithm

MEMD [41] is the generic extension of EMD [38], which is commonly used to analyze multi-channel data and flexibly deal with non-linear and non-stationary time series. MEMD effectively decomposes historical power peak load data and influential factors data into IMFs. There are different frequencies for different layers of the IMF. The first decomposed part is at the longest wavelength. The frequency decreases significantly as the number of decompositions increases. The last element of this computation is the Residue function. In case of multi-variable problems, local maxima and minima cannot be explicitly described [41]. Although the standard EMD algorithm can decompose a complex univariate load to a finite set of *imfs* and a residual, this algorithm presents several shortcomings in multi-dimensional datasets. The *imfs* extracted by using EMD from different time series do not necessarily correspond to the same frequency, and various load time series may extract different numbers of *imfs* components. From the computational cost view, it is challenging to match the differences *imfs* obtained from different time series. To conquer the inherent drawbacks of EMD, the MEMD technique is called upon in this paper to make the forecasting accuracy more reliable and drastically decrease the computational cost. The critical contribution of MEMD is to compute the local mean of n-dimensional signals by averaging all envelopes created by taking signal projections in different directions.

The four measured input variates for daily electricity load forecasting include the peak load, the valley load, the mean of the peak load, and the temperature were taken as input variables in this study. The MEMD can simultaneously decompose the m-variate inputs $X = \{X_1(t), X_2(t),..., X_m(t)\}$ ($t = 1,...,L$, where $L$ is the length of the time series) into n multivariate $imf(t)$ and a multivariate $r_n(t)$, where each component $imf_i(t)$ represents $\{imf_i^1(t), imf_i^2(t),..., imf_i^m(t)\}$ ($i = 1,...,n$) of the length $L$, and the residue $r_n(t)$ represents $\{r_n^1(t), r_n^2(t),..., r_n^m(t)\}$ of the length $L$.

The process of decomposition using MEMD is described briefly below:

1. The original multivariate signals $X(t) = \{X_1(t), X_2(t),..., X_m(t)\}$ are entered. The appropriate point set for the given input signals is generated with the Hammersley function.

2. The corresponding angles with the normalized Hammersley sequences are normalized in the range of 0 and $2\pi$. The direction vectors set $x^{\theta_i} = \{x_1^k, x_2^k,..., x_m^k\}, k = 1,..., K$ (where $K$ is the number of directions) corresponding to angles $\theta^k = \{\theta_1^k, \theta_2^k,..., \theta_{m-1}^k\}$ is established. The projection $\{p^{\theta_k}(t)\}_{k=1}^K$ of $X$ an along $kth$ direction is computed.

3. The extreme of the projection $\{p^{\theta_k}(t)\}_{k=1}^K$ is calculated on the instantaneous moment, $\{t_1^{\theta_k}\}_{k=1}^K$, where $1 \in [1, L]$ denotes the position of the extreme. The coordinates $(t_1^{\theta_k}, X(t_1^{\theta_k}))$ of the extreme dots are computed.

4. The spline interpolation is performed to obtain the multi-dimensional enveloped curves on the coordinates $(t_1^{\theta_k}, X(t_1^{\theta_k}))$ of the extreme dots. A mode would be excluded when the projected signal has inadequate extrema ($C$ denotes the number of extrema of the projected signal less than 3). The mean of the envelopes is calculated as follows.

$$\text{env\_mean}(t) = \frac{1}{K-C} \sum_{k=1}^{K-C} e^{\theta_k}(t) \quad (1)$$

5. The components $imf(t)$ are extracted in sequence from high to low frequency. As shown in Fig.1. Loop 1 is used to test whether $h(t)$ is $imf_i(t)$ obtained through the sifting criterion, and Loop 2 is used to determine whether $r(t)$ is the residue or is used to obtain next multivariate $imfs$ through the stopping criterion.

Through the above procedures of decomposition, the multi-dimensional signal $\{X(t)\}_{t=1}^{L}$ can be expressed as.

$$X(t) = \sum_{i=1}^{n} imf_i(t) + r_n(t) \quad (2)$$

Where n is the total number of $imfs$ obtained $i=1,...,n$ and $r_n$ is the residual.

**Table 2.** The daily history data of electricity load and temperature in NSW

| Date | Peak | Valley | Mean | Temperature |
|---|---|---|---|---|
| 2018/1/1 | 8747.49 | 5953.75 | 7365.99 | 26.4 |
| 2018/1/2 | 9395.79 | 5986.37 | 7895.12 | 24.8 |
| 2018/1/3 | 8235.15 | 6046.01 | 7501.37 | 23.0 |
| 2018/1/4 | 8651.08 | 5968.01 | 7557.75 | 22.6 |
| 2018/1/5 | 9727.00 | 6125.26 | 8021.31 | 24.1 |
| ... | ... | ... | ... | ... |
| ... | ... | ... | ... | ... |

To illustrate the MEMD result for the multivariate signal $\{X(t)\}$, as shown in Table 2, the daily history data of electricity load and temperature in NSW from Australia experimented in this study for peak load forecasting is listed. Let $X_t^P$, $X_t^V$, $X_t^M$ and $X_t^C$ be the daily peak load, the daily valley load, the daily mean of load, and the daily mean of temperature, respectively. The MEMD can decompose simultaneously the four-variate inputs $X(t) = \{X^P(t), X^V(t), X^M(t), X^C(t)\}, (t=1,...,L)$ into n multivariate time series $imf(t)$ and a multivariate time series $r_n(t)$, where each component $imf_i(t)$ represents $\{imf_i^P(t), imf_i^O(t), imf_i^M(t), imf_i^C(t)\}$ ($i=1,...,n$) of the length $L$ and the residue $r_n(t)$ represents $\{r_n^P(t), r_n^V(t), r_n^M(t), r_n^C(t)\}$ of the length $L$. The result of decomposition is shown in Fig. 2.

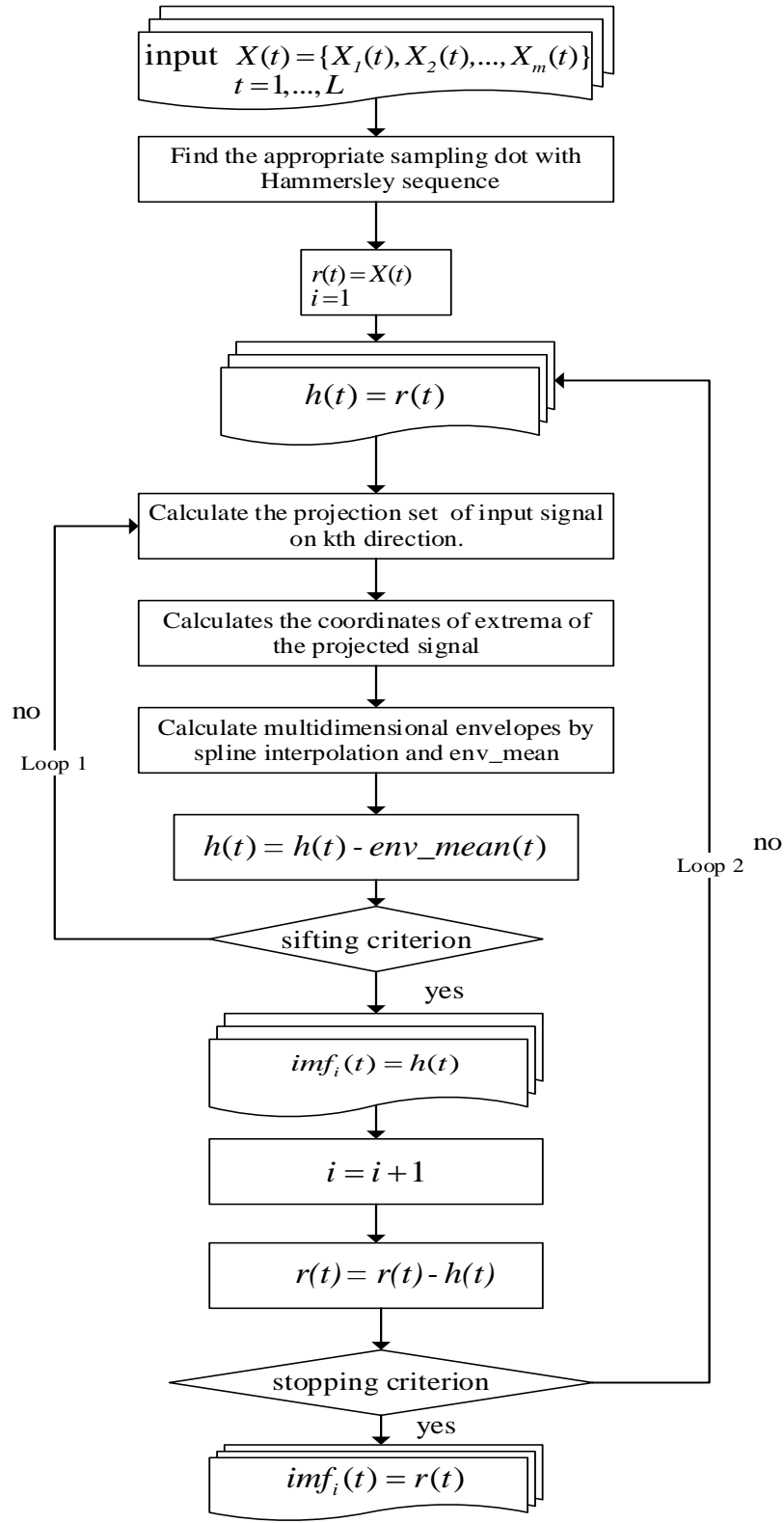

Fig. 1. The flowchart of the MEMD algorithm

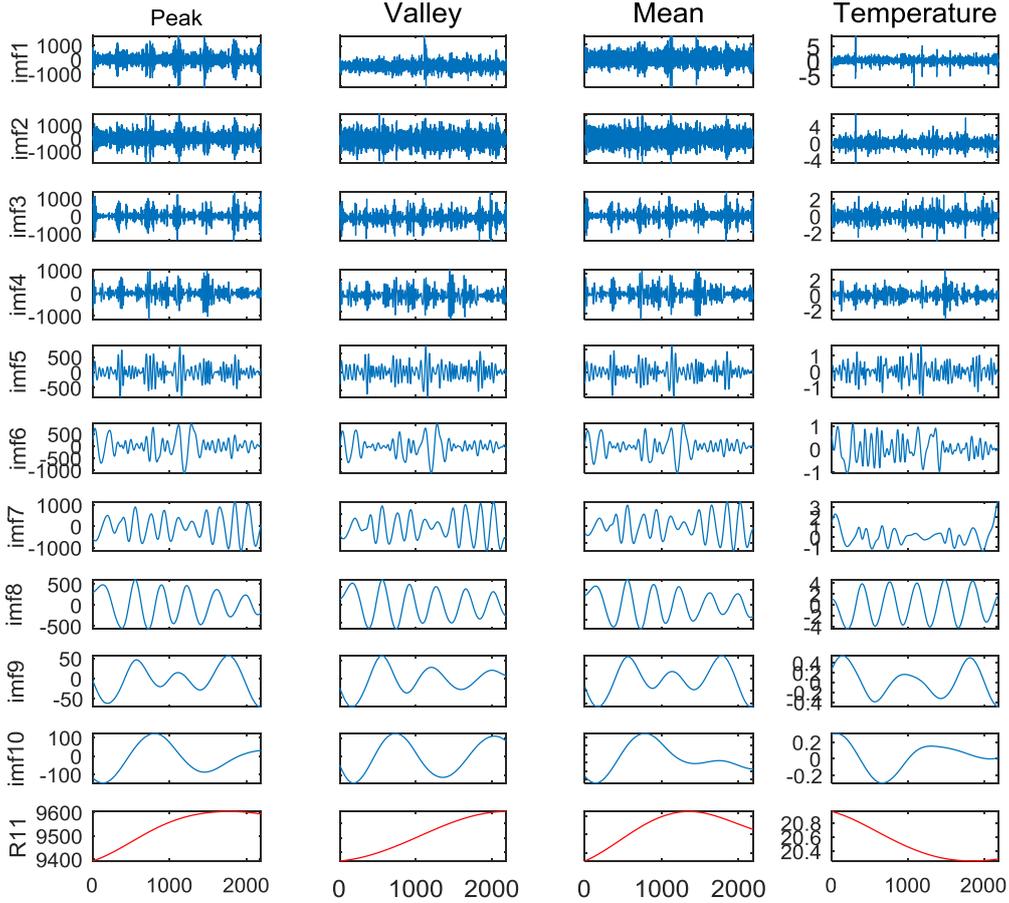

**Fig. 2.** The results of NSW using MEMD: *imfs* and *Rn*

## 2.2 Support Vector Regression (SVR)

SVR is typically used to model the non-linear and non-stationary regression problem [4]. To verify the performance of the proposed MEMD-based time-frequency analysis for day-ahead peak load forecasting, SVR is employed as a regression modeler due to its strong theoretical foundation and effective generalization ability. The preprocessing training samples are given as Eq. (3).

$$S = \{(X_i, Y_i)\} \subset R^m \times R, \quad i = 1, \text{L}, T \tag{3}$$

Where $X_i \in R^m$ is the $ith$ m-dimensional input vector, $Y_i \in R$ is the corresponding actual target load, and $T$ is the number of training samples. SVR is employed to generate the mapping model $f: R^m \to R$ from the multivariate input spaces $R^m$ to output space $R$. $\varepsilon$. Based on Vapnik's main idea, SVR first applies a non-linear function $\varphi(\cdot)$ to map the input spaces $X_i$ to a high-dimensional feature space $\varphi(X_i)$, then performs linear regression using the value of $\varphi(X_i)$. The linear regression function is $f(X) = w^T \varphi(X) + b$, Where $f(X)$ is the estimated value. To ensure the flatness of the linear regression function, we must find the smallest $w^T$. The linear regression problem can be transformed into solving the following optimization problem as Eq. (4). The restricted condition is Eq. (5).

$$\min \quad \frac{1}{2}\|w\|^2 \qquad (4)$$

$$\left|Y_i - (w^T \varphi(X_i) + b)\right| \leq \varepsilon \qquad (5)$$

Where, $\varepsilon$ is the width of an insensitive loss function. Besides, considering the possible error, two nonnegative variables $\xi_i, \xi_i^*$ introduced are the slack variables. The optimization problem of model (4) and (5) can be transformed into model (6).

$$\min \quad \frac{1}{2}\|w\|^2 + C\sum_{i=1}^{T}(\xi_i + \xi_i^*)$$

$$s.t. \quad w^T \varphi(X_i) + b - Y_i \leq \varepsilon + \xi_i^*, (i=1,...,T) \qquad (6)$$
$$\qquad Y_i - (w^T \varphi(X_i) + b) \leq \varepsilon + \xi_i, (i=1,...,T)$$

Where, $C$ is a penalty parameter. By employing Lagrangian function and KKT conditions for optimality, the final solution of the primal problem can be represented in the following form:

$$w = \sum_{i=1}^{T}(\alpha_i - \alpha_i^*)\varphi(X_i) \qquad (7)$$

$$f(X) = \sum_{i=1}^{T}(\alpha_i - \alpha_i^*) \cdot K(X_i, X_j) + b \qquad (8)$$

Where $\alpha_i$ and $\alpha_i^*$ are nonnegative Lagrange multipliers, and $K(X_i, X_j)$ equal to $\varphi(X_i) \cdot \varphi(X_j)$ is a kernel function and can simplify the mapping procedure. It is important to note that the implementation of radial basis function (RBF) can reduce feature dimensions and re-rank the features. Due to its superior performance, RBF, $K(x_i, x_j) = \exp\left(-\gamma \|x_i - x_j\|^2\right), \gamma > 0$, is selected as the kernel function in this study, considering both the prediction accuracy of the non-linear model and the number of the hyper-parameters in SVR.

## 2.3 Particle Swarm Optimization (PSO)

To ensure the forecasting accuracy, we need to carefully tune three hyper-parameters of SVR, namely the penalty parameter $C$, the width of insensitive loss function $\varepsilon$, and the RBF kernel parameter $\gamma$, based on the training dataset. According to the literature mentioned above, experimental results have demonstrated that the PSO algorithm can solve the hyper-parameter fine-tuning of SVR with global search capability remarkably.

When exploring the optimal solution in the search space of the PSO parameters, each particle represents a potential solution to parameters optimization of SVR. The parameters of PSO selected in a trial-error fashion are presented in Table 3.

**Table 3.** The final parameters of PSO

| Parameters | Abbreviation | Value |
|---|---|---|
| Swarm size | pop | 30 |
| Number of iterations | N | 100 |
| Cognitive coefficients | c1 | 1.5 |
| Interaction coefficients | c2 | 1.6 |
| Initial inertia weight ($w$) | wMax | 0.9 |
| Final inertia weight ($w$) | wMin | 0.4 |

The status of each particle depends on its position ($X$) and velocity ($V$). In this study, the position and velocity of the m-dimensional particle i can be denoted as following at the $n$th iteration:

$$X_i^n = \{X_{i1}^n,...,X_{im}^n\}, n=1,...,N \quad (9)$$

$$V_i^n = \{V_{i1}^n,...,V_{im}^n\}, n=1,...,N \quad (10)$$

The optimization procedure of running PSO is described briefly below:

(1) Initialize the PSO parameters as in table 3 and randomly generate the initial position and velocity of each particle following Eq. (11) and (12). Where $r$ is a random number in the range of [0, 1], $d$ is the dimension of particle $i$. $X_{max,d}$, and $X_{min,d}$ denote the maximum boundary value and the minimum boundary value of the position of each particle of the $dth$ parameter to avoid the particle converging slowly, respectively. $V_{max,d}$ and $V_{min,d}$ are often used to constraint the velocity of each particle of the $dth$ parameter to hinder the particle flying outside the search space, respectively.

$$X_{i,d} = X_{min,d} + r(X_{max,d} - X_{min,d}) \qquad (11)$$

$$V_{i,d} = V_{min,d} + r(V_{max,d} - V_{min,d}) \qquad (12)$$

(2) Calculate the fitness of each particle. The mean absolute percentage error (MAPE) is employed as the fitness function in this study.

(3) Search for the optimal fitness value, namely the minimum of MAPE. The fitness function of MAPE is used to evaluate the superiority of each particle. When the smaller value of MAPE in $i$th iteration the local optimal solution and the global optimal solution are judged whether to be updated. The optimal fitness of each particle itself ( $pbest$ ) denoting the local optimal solution and the optimal fitness of all particles ( $gbest$ ) indicating the global optimal solution are found.

(4) Each particle dynamically adjusts its flight status according to its own flight experience ( $pbest$ ) and companion's flight experience ( $gbest$ ) as Eq.(13) and Eq.(14).

$$V_{id}^{t+1} = w \times V_{id}^{t} + c_1 \times r_1 \times (pbest_{id} - x_{id}^{t}) + c_2 \times r_2 \times (gbest_{id} - x_{id}^{t}) \qquad (13)$$

$$x_{id}^{t+1} = x_{id}^{t} + V_{id}^{t+1} \qquad (14)$$

Where $w$ is inertia weight that is equal to wMax-((wMax-wMin)/N)*$t$ ($t$ denotes the time of current iteration)

(5) The optimal solution is found by iterations. If the termination condition is satisfied, the iteration stops. Otherwise, the loop returns to step 2.

Three parameters of SVR corresponding to the global optimal particles are used to train the SVR prediction model.

**2.4 The proposed MEMD-PSO-SVR modeling framework**

According to the methods mentioned above, the overall flow of the proposed MEMD-PSO-SVR modeling framework is elaborated in detail.

There are three main objectives for peak load forecasting in this study:

- First, is to simultaneously preprocess the multi-dimensional time series with a time-frequency analysis, which can remarkably improve future electricity peak load by extending the univariate scenarios.

- Secondly, to avoid over-fitting along the training of the developed model, decreasing the computational complexity of modeling and forecasting with good generalization ability.

- Finally, to explore the optimal modeling parameters, ensuring to obtain the optimal solution of electricity peak load forecasting.

As shown in Fig. 3, the specific studies of the proposed MEMD-based model are as follows:

*Step 1* : The multivariate channels $X = \{peak, valley, mean, Temperature\}$ is first decomposed simultaneously into $n$ multivariate *imfs* extracted and a multivariate $r_n$ using the MEMD approach. In order to boost the accuracy, time-domain analysis can assess the possible association of the impact factors of electricity load patterns, and frequency-domain analysis can provide a more detailed exploration of the operations inherent in the data sets.

The time-domain analysis is performed on the original data set to capture electricity load patterns, as shown in Fig. 1. MEMD with time-frequency analysis is employed to decompose the time-domain signal simultaneously into finite various frequencies waves and convert these random signals into stable and predictable signals of different frequencies with effectively extracting inherent information among relevant variables for significantly improving the accuracy of prediction.

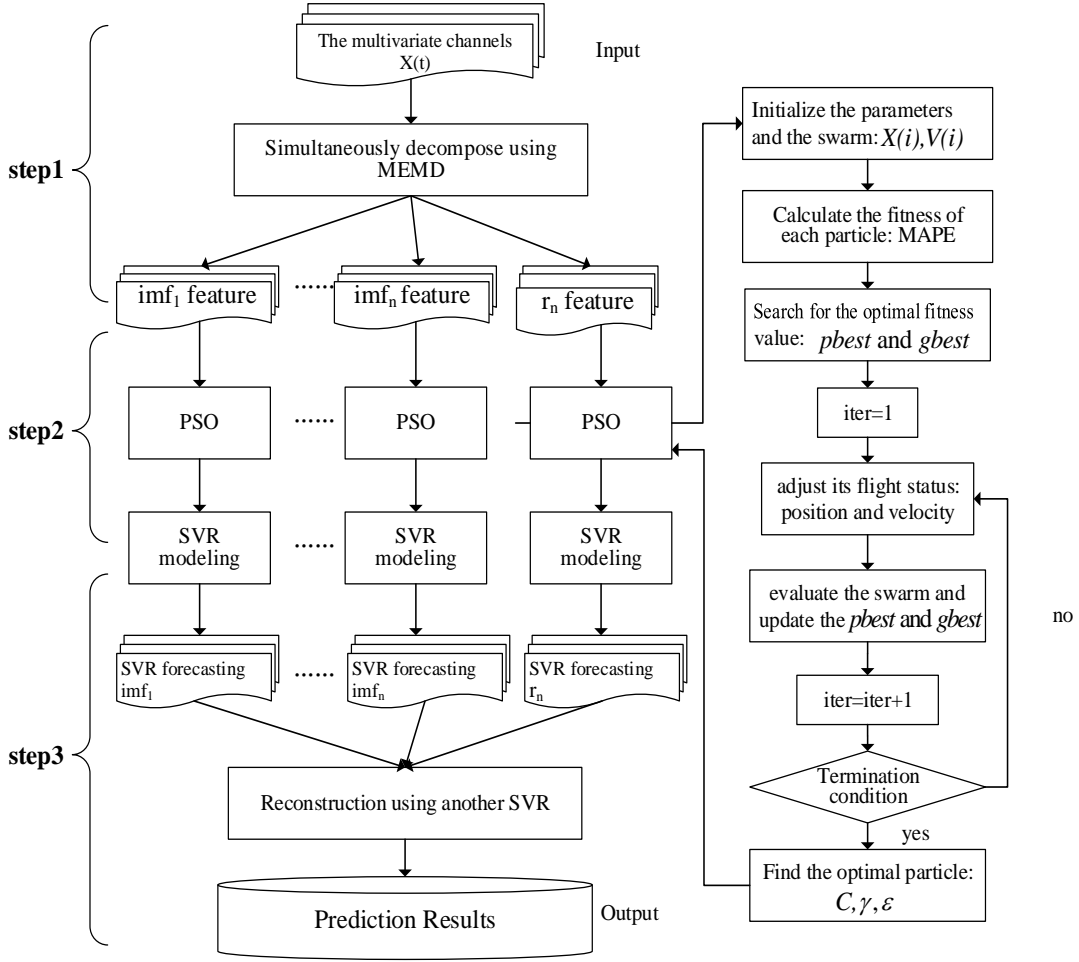

Fig. 3. The MEMD-PSO-SVR modeling framework

*Step 2*: The hyper-parameters of SVR are optimized by employing PSO. To ensure the forecasting accuracy of SVR, three SVR parameters, namely $C$, $\varepsilon$ and $\gamma$, are tuned carefully by MAPE in training sets. The components of *imfs* and $r_n$ extracted from the original signal $X$ are inputted to explore the optimal parameters of SVR before modeling and forecasting each component, respectively, by SVR. PSO is a stochastic optimization algorithm based on swarm intelligence. Each possible solution of PSO is defined as a particle of population, and the fitness function of MAPE is used to evaluate each particle's superiority. Each particle dynamically adjusts its flight velocity and position according to its cognition and social learning to search for the optimal solution, that is, the global optimal solution. The parameters of SVR corresponding to the global optimal solution are used to establish the SVR prediction model.

*Step* 3: SVR is employed to establish the model and predict each component extracted using the MEMD technique in this study. PSO has explored the optimal hyper-parameters of SVR. Therefore, the corresponding improved forecast of each component extracted from *Step 1* can be obtained, which needs to be reconstructed to obtain the final electricity peak load forecast. Another SVR is used to rebuild as an ensemble tool according to the projections of each component. The ensemble result would be considered as the final forecast of the peak load of the next day.

In a few words, the multivariate channels are first inputted into the proposed MEMD-PSO-SVR model to be decomposed simultaneously using MEMD technique. Then three hyper-parameters of SVR are optimized by PSO. Finally, PSO-based SVR is employed to establish a model and predict each component respectively extracted using MEMD technique. The forecast of each component is integrated to obtain the final prediction.

To verify the proposed MEMD-PSO-SVR modeling framework's performance, we examine the effectiveness of the proposed hybrid model against six other comparable models, including traditional SVR, conventional BPNN, EEMD-SVR, EEMD-PSO-SVR, MEMD-SVR, and MEMD-PSO-BPNN. The implementations of these methodologies are presented in the next section.

## 3. Research design

### 3.1 Data sets

Two real-world load data sets from the New South Wales (NSW) and the Victoria (VIC) in Australia [54] have been considered to verify the superiority of the proposed MEMD-PSO-SVR hybrid model, respectively. The main reason of adopting the samples is that NSW and VIC are the two largest states in Australia, and the two states are recognized as representatives of comprehensive electricity load of power system, including the electricity consumption of industrial, agricultural, post and telecommunications, transportation, municipal, commercial and urban-rural residents. Simulation experiment data sets use the first two-thirds of daily historical data of two states as the training samples to establish a model to forecast the peak load of the rest. That is to say that the samples are divided into the training datasets and the testing datasets. The two daily datasets include the peak load, the valley load, the mean load and the corresponding temperature over the same period. Table 4 presents experimental data distribution of the two data sets.

Table 4. Period and size of experimental data sets

| Sample data | The period of the time series | sample size | training set | testing set |
|---|---|---|---|---|
| NSW | JAN. 1, 2014 to Dec 31, 2019 | 2191 | 1460 | 734 |
| VIC | JAN. 1, 2015 to Dec 31, 2019 | 1826 | 1217 | 609 |

For the two multi-dimensional original series, let $x^P$, $x^V$, $x^M$, and $x^C$ denote the peak load, the valley load, the mean load, and the temperature, respectively, and let $\hat{x}^P$ denote the predicted value of the peak load. As such, a multi-dimensional vector $x_t = \{x_t^P, x_t^V, x_t^M, x_t^C\}, (t=1,...,T)$ composed by the influencing factors represents the input on the $t$th day and $x_{t+1}^P$ represents the output on the next day. Each dataset is represented as a set $D = \{(X_i, Y_i) \in (R^{4d} \times R)\}_{i=d}^T$ of the proposed model, where $X_i = [x_t^P, x_t^V, x_t^M, x_t^C, ..., x_{t-d+1}^P, x_{t-d+1}^V, x_{t-d+1}^M, x_{t-d+1}^C]$ denoting the lagged values of a historical data are set as inputs and $Y_i = x_{t+1}^P$ is set as output of $X_i$. The $T-d+1$ input-output pairs are constructed for modeling and forecasting through $X_i$ and $Y_i$. Where the embedded dimension, $d$, is set following a fashion of trial and error. Details can be found in subsection 3.3.

**3.2 Evaluation metrics**

To measure the forecasting accuracy of the proposed MEMD-PSO-SVR hybrid model, four statistical errors, namely R-squared ($R^2$), root mean square error (RMSE), mean absolute percentage error (MAPE) and directional accuracy (DA), are selected to demonstrate the prediction accuracy. The four evaluation metrics can provide reference for decision-making of power system. $R^2$ is formulated as Eq. (15), which is employed commonly to evaluate the fitting level of different models on the same testing set. The larger the value of $R^2$ is, the better the performance of forecasting for model is[12, 55]. RMSE is formulated as Eq. (16), which is employed commonly to compare the squared absolute error between the real load and the predicted load. The smaller the value of RMSE is, the closer the predicted load gets to the real load [9, 56, 57]. MAPE is formulated as Eq. (17), which is employed usually to evaluate the relative percentage error between the real load and the predicted load. The smaller the value of MAPE is, the better the forecasting performance of model is[9, 28, 57, 58]. DA is formulated as Eq. (18), which is an approach to measure the validity of the forecasting direction and provides the moving trend for investors[59].

$$R^2 = 1 - \frac{\sum_{t=1}^{N}(Y_t - \hat{Y}_t)^2}{\sum_{t=1}^{N}(Y_t - \bar{Y})^2} \tag{15}$$

$$RMSE = \sqrt{\frac{\sum_{t=1}^{N}(Y_t - \hat{Y}_t)^2}{N}} \tag{16}$$

$$MAPE = \frac{1}{N}\sum_{t=1}^{N}\left|\frac{Y_t - \hat{Y}_t}{Y_t}\right| \times 100 \tag{17}$$

$$DA = \frac{\sum_{t=2}^{N} d_t}{N-1} \times 100 \ , t = 2,...,N$$
$$s.t. \ d_t = \begin{cases} 1, & \text{if } (y_t - y_{t-1})(\hat{y}_t - \hat{y}_{t-1}) \geq 0 \\ 0, & \text{otherwise} \end{cases} \tag{18}$$

Where $N$ represents the length of the testing dataset. $Y_t$ and $\hat{Y}_t$ are the real peak load and the predicted peak load at the $t$th point-in-time respectively. $\bar{Y}$ is the mean of the real peak load, and $d_t$ denotes the direction of the forecast and the observation. Moreover, to demonstrate the proposed MEMD-PSO-SVR hybrid model's forecasting reliability in comparison to other competing models, two statistical test methods are used to evaluate the models with one another to find a significant difference in the testing datasets. Firstly, the Analysis of Variance (ANOVA) test is a preliminary test, which is a powerful approach of testing the null hypothesis of means equality[60], to determine whether an obvious performance discrepancy could be found among all the comparative models. Secondly, the Diebold-Mariano (DM) [61] test is employed to assess the significant difference of prediction accuracy between the proposed hybrid model and other comparative models at a certain significance level as Eq.(19). However, it is worth noting that the DM test can effectively eliminate the constraint of stochastic difference of samples，that utilized to determine whether or not each model's predicting errors are reduced when compared to the others, and can provide reliability for the comprehensive evaluation.

$$DM = \frac{d_{mean}}{d_{std}}$$

$$\text{s.t.} \begin{cases} E_a = [a^1, a^2, ..., a^T] \\ E_b = [b^1, b^2, ..., b^T] \\ d^i = a^i - b^i \\ d_{mean} = \frac{\sum_{i=1}^{T} d^i}{T} \\ d_{std} = \sqrt{\frac{\sum_{i=1}^{T} (d^i - d_{mean})^2}{T-1}} \\ i = 1, ..., T \end{cases} \quad (19)$$

### 3.3 Experimental implementations

To verify the superiority of the proposed hybrid MEMD-PSO-SVR model, seven models, including six comparative models and the proposed model, are implemented with the two real-world datasets. These models are established from three tasks: time-frequency decomposition, multivariate considering the temperature, and parameters optimization. Back propagation neural network (BPNN) is chosen as a fashionable neural network representative to compare with SVR. The benchmark models for SVR and BPNN are used as single-model representatives of multivariate inputs compared with other hybrid models. EMD-based SVR is as a univariate representative without considering the important factor of temperature compared with MEMD-based hybrid models. MEMD-based benchmark model for SVR is as a multivariate decomposition representative without optimizing parameters of SVR. The MEMD-PSO-BPNN and the proposed hybrid models are employed to perform one-step ahead peak load forecasting with parameter optimization and time-frequency decomposition. The detailed implementations of these methods are introduced as follow.

The proposed MEMD-PSO-SVR hybrid model is performed in the environment of MATLAB R2016b, where MEMD plays an extremely important role in improving the forecasting accuracy. In the function of MEMD, 'num_direction' is an optional parameter to specify the number of projection directions for input variables, which is set to 256 in this study. However, it is noted that it should far exceed the dimensionality of input variables. The PSO algorithm is employed to search for the optimal parameters of SVR for further improving the performance of prediction. The parameters of PSO are selected in a trial-error fashion as Table 3. SVR is implemented by LIBSVM(Version 3.24), which is a library for Support Vector Machines provided by Chang and Lin [62]. To ensure the forecasting accuracy, three hyper-parameters fine-tuning of SVR, namely $C$, $\varepsilon$, $\gamma$, were determined by PSO based hyperparameter optimization in the training process. The search space of parameters is defined: $C \in [10^{-1}, 10^3]$, $\gamma \in [10^{-3}, 10^3]$, and $\varepsilon \in [10^{-3}, 10^{-1}]$, respectively. To generate and evaluate the optimal parameters in SVR, the average of MAPE is selected as the fitness function of PSO. The smaller the value of MAPE is, the better the particle is for modeling and prediction. During the training process, the potential size of the embedded dimension is set from one to sixteen. We need to trade off the prediction accuracy and the computational time for the optimal embedded dimension in two real-world samples.

Consequently, the best one $d = 6$ is selected, as shown in table 5 and Fig. 4.

Table 5
Embedded dimension in term of the performance through trial-and-error

| Embedded dimension | MAPE/% | | Run time (second) | |
| --- | --- | --- | --- | --- |
| | NSW | VIC | NSW | VIC |
| 1 | 3.1082 | 5.0634 | 13199.1741 | 12105.2268 |
| 2 | 2.5129 | 3.3417 | 14333.6772 | 13213.2485 |
| 3 | 2.0967 | 2.8726 | 16408.4077 | 14243.5314 |
| 4 | 1.5234 | 2.0420 | 17501.3427 | 15286.5385 |
| 5 | 1.2777 | 1.6031 | 18511.5306 | 15624.1843 |
| **6** | **0.7866** | **1.0783** | **18729.0915** | **16169.2936** |
| 7 | 0.8592 | 1.2136 | 19558.1864 | 16348.3070 |
| 8 | 0.9907 | 1.3087 | 21603.9049 | 18362.1964 |
| 9 | 1.0839 | 1.5008 | 21660.9941 | 18397.4960 |
| 10 | 1.1787 | 1.8137 | 22683.1256 | 19401.1378 |
| 11 | 1.3084 | 2.1156 | 23723.3251 | 19430.3383 |
| 12 | 1.5516 | 2.3131 | 24743.0479 | 21453.9986 |
| 13 | 1.7621 | 2.5731 | 26799.6080 | 23468.4799 |
| 14 | 1.8333 | 2.7864 | 29840.7757 | 25483.4408 |
| 15 | 2.3101 | 3.1659 | 32867.1803 | 28510.2463 |
| 16 | 2.4449 | 3.3199 | 35930.6161 | 30521.1229 |

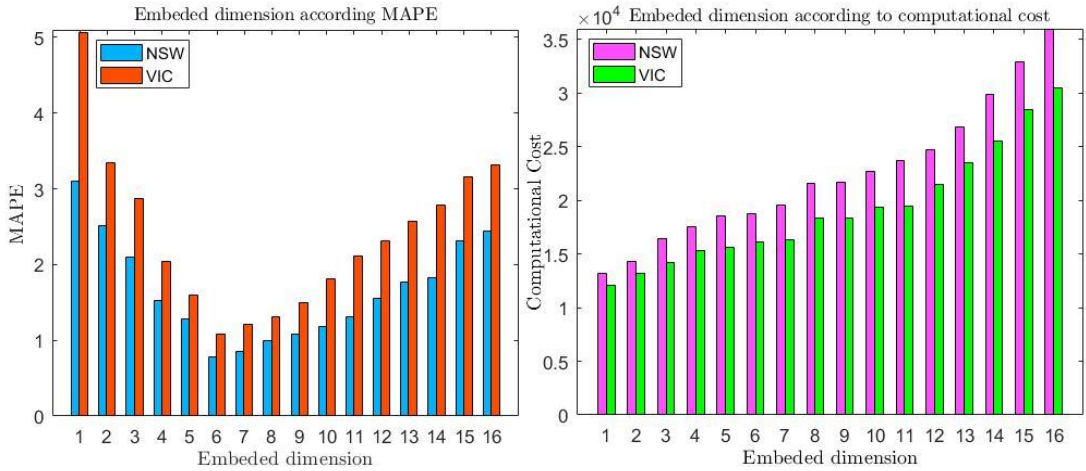

Fig. 4. Tradeoff prediction accuracy and computational cost

The traditional EMD-SVR model is also implemented in the same environment as other models. However, it is noted that the ensemble EMD(EEMD) algorithm provided by Wu and Huang [63] is employed to decompose the single variable of the peak load ignoring the temperature. In the program of EEMD, 'Nstd' denotes the ratio of the standard deviation of the added noise and that of the original signal, which is set to zero. NE denotes the ensemble number for the EEMD, which is set to 1. The three hyper-parameters (C,$\gamma$,$\varepsilon$) in SVR are set to 64, 2, and 0.001, respectively, by the rule of thumb.

In the single SVR model, the parameters selection of SVR is the same as the EMD-SVR model mentioned above. The input variables and output variable are the same as the proposed hybrid model. In the hybrid MEMD-SVR model, the parameters of MEMD are set as same as that of the proposed model. The parameters selection of SVR is the same as the EMD-SVR model mentioned above.

The hybrid MEMD-PSO-BPNN model is performed in the environment of Matlab R2016b, The parameters of MEMD are set as same as that of the proposed model. BPNN, as a representative of neural network, is adopted using the MATLAB NNET toolbox. Input layer is set to 4, output layer is set to 1.The number of hidden layers is set at 10 by testing from8 to 20, and the optimal number of hidden nodes that minimizes the error rate on the training set is determined. In the training phase, gradient descent with momentum algorithms is applied to update weight and bias values. In the single BPNN model, the parameters of BPNN are set the same as the MEMD-PSO-BPNN model mentioned above. To guarantee the reliability of comparison, the average performances of each model are used on comparison in testing set.

## 4. Experimental results

### 4.1 Experimental data analysis

Half-hourly historical electricity load data were collected form NSW and VIC in Australia [54]. The daily peak load, the valley load, the mean load along with the daily average temperature load were extracted and aggregated to generate the experimental dataset. Fig.5(a) and Fig.5(b), indicate the presence of non-linearity and non-stationarity as the seasonal change characteristics and significant fluctuation of the wave form for the peak load, the valley load, and the mean of load, respectively. It is not difficult to find that the curve of peak load has highly stochastic fluctuation. As discussed above, the influential factors on daily electricity demand include meteorology, season, time segments, holidays, load lags, load distribution, as well as social, political and economic factors. To avoid accumulative error and reduce hybrid models' computational complexity, temperature is only considered as an exogenous variable in this paper.

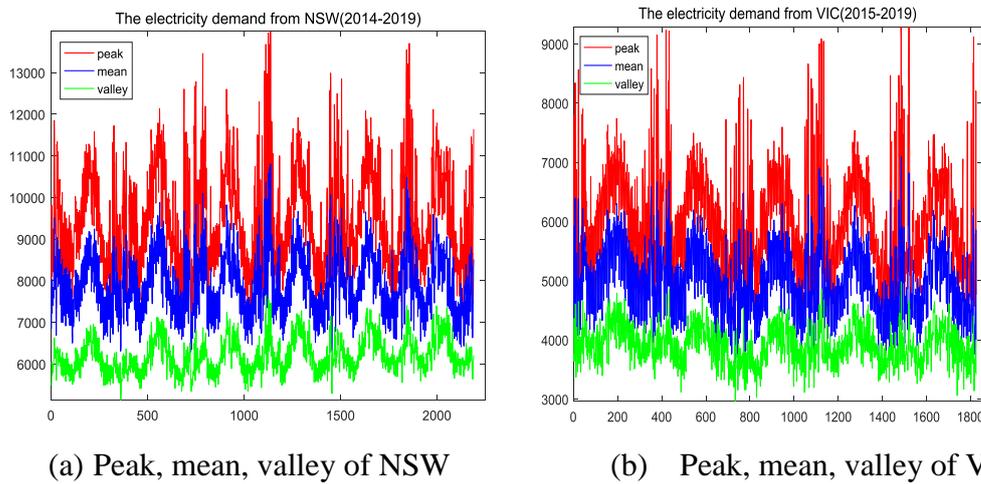

(a) Peak, mean, valley of NSW  (b)  Peak, mean, valley of VIC
Fig. 5. Electrical demand of each historical load: peak, mean, valley

Fig.6(a) and Fig.6(b), illustrate the daily mean of historical load is selected to verify the correlation with the daily mean of temperature. These two figures (Fig. 6(a) and Fig. 6(b)) demonstrated the temperature is the lowest or highest, the mean of power load reaches the peak load. The relationship between the mean of load demand and the mean of temperature has verified the importance of temperature for the day-ahead load forecasting. Therefore, the historical temperatures are worthwhile to select as the input variables.

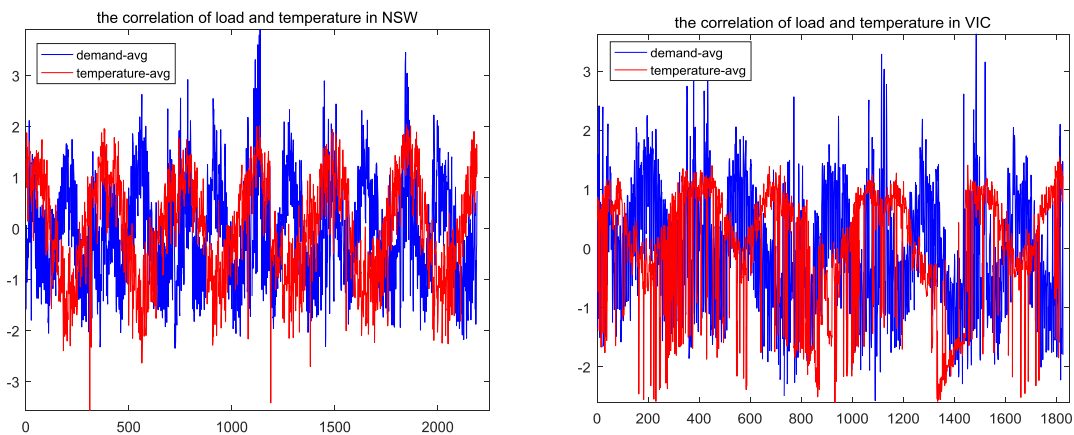

(a) Mean and temperature of NSW　　　　(b)　Mean and temperature of VIC

Fig. 6. The correlation of load and temperature: mean and temperature

**4.2 Decomposition analysis**

To predict more accurately the day-ahead peak load and find the evolution intrinsic pattern of fluctuation of electrical demand, the multivariate input variables of each dataset are simultaneously decomposed using the MEMD algorithm. As shown in Fig.2 and Fig. 7, the results of decomposition are ten multivariate *imfs* components in the frequency from high to low and one multivariate residue $Rn$ component. However, it should be noted that the number of multivariate *imfs* components extracted would be determined by the size of samples. If there is a great deviation or an integral multiple of 2 in the size of the sample collected from NSW and VIC in Australia, the number of the components extracted is different.

Furthermore, it is not difficult for the time-frequency spectrum observed to find that all multivariate *imfs* components fluctuate up and down at zero, however, the waveform of *imfs* is still asymmetric. To further verify the evolution intrinsic pattern of fluctuation of electrical demand, we reconstruct the components of each channel.

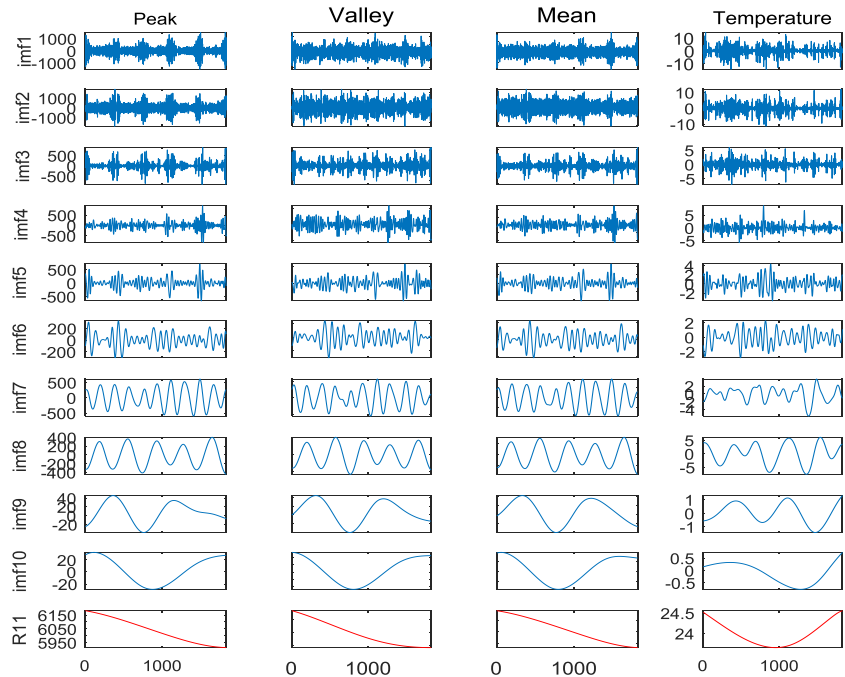

**Fig. 7.** The results of VIC using MEMD: $imfs$ and $Rn$

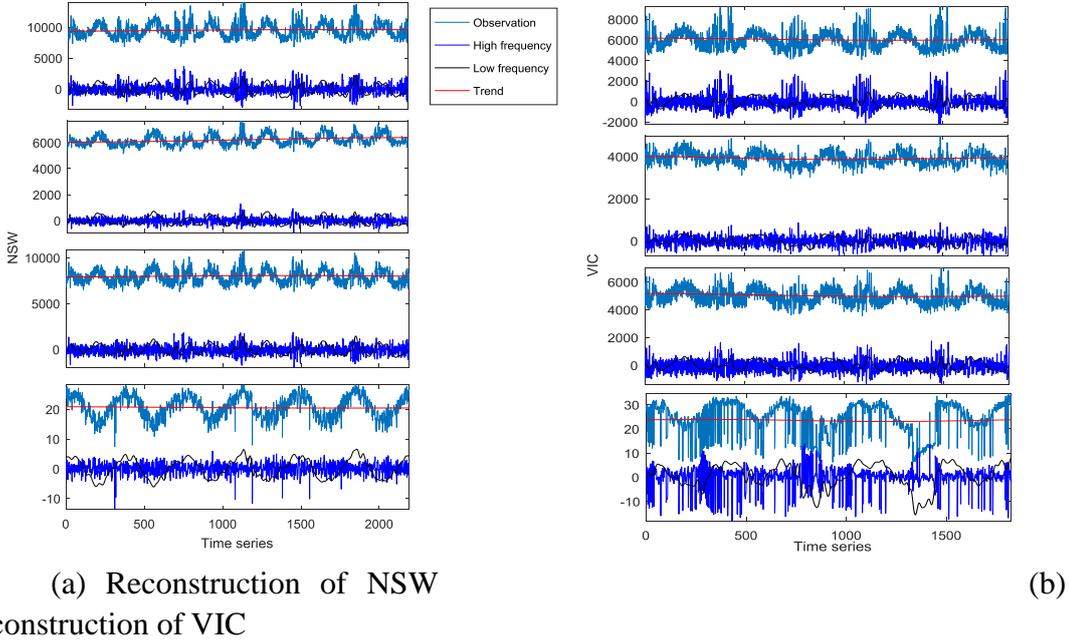

(a) Reconstruction of NSW  (b) Reconstruction of VIC

Fig.8. Reconstruction of $imfs$ and the residue $Rn$

As shown in Fig.8(a) and Fig.8(b), The observation of original signal depends mainly on the residue $Rn$ represented the main trend of each channel, which is not affected by the factors of the corresponding $imfs$ components. The prediction of the residue $Rn$ is extremely crucial for decision-makers of long-term load forecasting, while high-frequency components ($imf_1 \sim imf_5$) fluctuated with mini-amplitude reflect the normal fluctuation item in peak load forecasting, such as workday, weekend, electrical demand of residents and enterprise's manufacturing. Although the high-frequency components have mini-amplitude compared with the main trend of electrical demand, the peak load forecast is also very important for the production safety and human life such as cooling, warming, as well as industrial operations. Moreover, low frequency ($imf_6 \sim imf_{10}$) components implied important factors, such as social activities and seasonality. Although the frequency of the components (time series load) is very low, it has a great influence on the electrical demand. However, some unavoidable random factors would affect the accuracy and the computational expense of prediction.

## 4.3 Statistical analysis

All models are implemented on the two real-world testing sets. Afterwards, the DA, MAPE, RMSE and $R^2$ are calculated for the testing set. We repeatedly perform the seven models for 50 times, which generated 50 results of DA, MAPE, RMSE and $R^2$ for each model. The performance is ranked in terms of the mean of DA, MAPE, RMSE and $R^2$ of the 50 times in NSW and VIC. Prediction accuracy measure of different models for the two testing datasets are presented in details in Table 6.

Table 6 shows the comparisons of the mean difference of DA, MAPE, RMSE and $R^2$ of all models across various datasets. In all models, the single BPNN model has the worst performance on DA and $R^2$. The reason might be the intrinsic limitations of the single BPNN model for peak load forecasting, such as over-fitting on training sets and local optimum. Comparing the classical EEMD-SVR model and the MEMD-SVR models, the EEMD-SVR model achieves lower accuracy because of not considering the other relevant covariates, such as the temperature. However, because of the time-frequency decomposition of EEMD or MEMD, hybrid models MEMD-SVR and EEMD-SVR obtain better performance than single model. Furthermore, MEMD-PSO-BPNN has good performance on DA and $R^2$, which is due to the optimization technique of PSO and the decomposition algorithm of MEMD. Comparing the MEMD-SVR model, the proposed MEMD-PSO-SVR model is more accurate by using the optimization of PSO. For the analysis of error measures above, the proposed hybrid model was achieved higher performance than other examined models. Error measures with the best performance are highlighted in bold in Table 6. Furthermore, the EEMD-PSO-SVR modeling framework proposed in Ref.[64], which dealt with univariate short-term load forecasting based on SVR optimized using PSO, is compared with the proposed model of this paper for emphasizing mainly the importance of exogenous variable and the effectiveness of multivariate decomposition.

To further verify the performance of prediction, two statistical methods (ANOVA test[60] and DM test[61]) are employed to determine whether a statistically significant difference exist among the examined models for peak load forecasting. The ANOVA test is initially used to determine if the proposed MEMD-PSO-SVR hybrid model exhibits a significant performance difference when compared to the other examined models. However, for comparison between all possible pair wise models could not be found, if the null hypothesis of this ANOVA test is rejected, the DM testis then performed to identify all pair-wise differences simultaneously at the 0.05 level. Note that the DM test can effectively eliminate the constraint of random sampling dynamics and provide comprehensive evaluation for accuracy and stability. The DM test results in the testing sets are shown in Table 7.

The forecasting errors of seven different hybrid models were tested using the DM test, and the DM test results indicated that the MEMD-PSO-SVR test result consistently exceed the 5% significance level upper bound in all cases. Table 7 presents the statistical results with the corresponding P values, which reveals that the proposed forecasting hybrid model (MEMD-PSO-SVR) reaches superior performance to the other hybrid models at the 1% significance level. According to the comprehensively evaluation of DM test, it can be reasonably concluded that the proposed modeling framework not only has higher forecasting accuracy than other models, but also shows significant differences at a certain significant level, which further verifies the performance robustness of the proposed model in peak load forecasting.

**Table 6** The forecast accuracy of seven models in the two real-world testing sets

| Model | Input variable | | DA/% | | MAPE/% | | RMSE | | $R^2$ | | Run time (second) | |
|---|---|---|---|---|---|---|---|---|---|---|---|---|
| | History load | Temperature | NSW | VIC | NSW | VIC | NSW | VIC | NSW | VIC | NSW | VIC |
| SVR | added | added | 72.31 | 68.89 | 4.365 | 5.496 | 443.3 | 532.0 | 0.750 | 0.743 | 1.02 | 0.95 |
| BPNN | added | added | 63.31 | 65.89 | 4.267 | 4.996 | 475.3 | 513.4 | 0.679 | 0.651 | 1.32 | 1.15 |
| EEMD-SVR | added | no | 82.40 | 78.66 | 2.525 | 3.047 | 351.1 | 346.4 | 0.906 | 0.829 | 19.46 | 13.73 |
| EEMD-PSO-SVR | added | no | 83.17 | 80.21 | 2.502 | 2.989 | 253.5 | 335.7 | 0.912 | 0.851 | 72028 | 63530 |
| MEMD-SVR | added | added | 86.76 | 83.43 | 2.495 | 2.946 | 183.4 | 351.1 | 0.924 | 0.864 | 186.2 | 104.7 |
| MEMD-PSO-BPN | added | added | 88.13 | 83.95 | 4.213 | 4.842 | 345.2 | 431.6 | 0.869 | 0.826 | 31620 | 21705 |
| MEMD-PSO-SVR | added | added | **92.76** | **90.43** | **0.786** | **1.078** | **113.9** | **153.7** | **0.954** | **0.924** | 18729 | 16169 |

NB: The values in bold denoted the best value is significant among the seven models at the 0.05 level.

**Table 7**
The DM test values between the proposed model and other six models in the two real-world testing sets

| Model | NSW | | VIC | |
|---|---|---|---|---|
| | DM | P-value | DM | P-value |
| SVR | 6.8125 | 0.0000* | 7.6345 | 0.0000* |
| BPNN | 9.3244 | 0.0000* | 9.8127 | 0.0000* |
| EEMD-SVR | 5.6834 | 0.0000 * | 6.7852 | 0.0000* |
| EEMD-PSO-SVR | 5.2951 | 0.0000* | 5.8917 | 0.0000* |
| MEMD-SVR | 4.5726 | 0.0000* | 5.3281 | 0.0000* |
| MEMD-PSO-BPNN | 6.2731 | 0.0000* | 7.0458 | 0.0000* |

Note: * denotes the significance levels are 99%.

The snapshots of prediction results with the actual value in both training sets and testing sets in NSW and VIC are shown in Fig. 9-10.

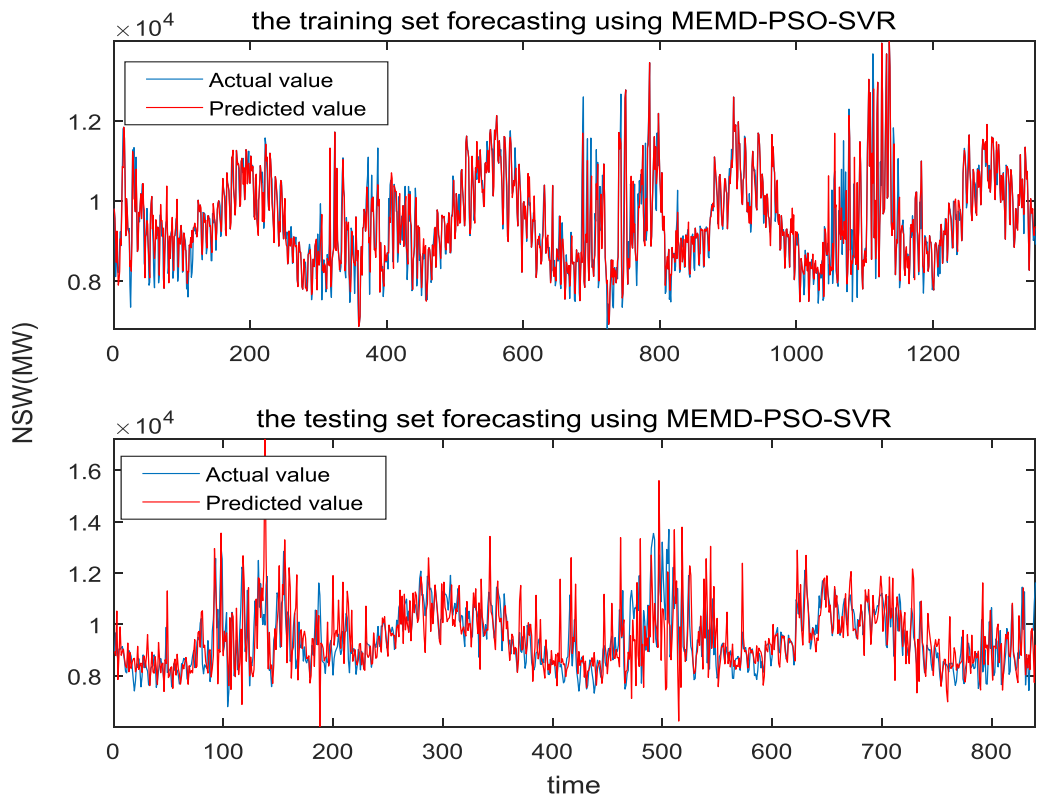

Fig. 9. The prediction of NSW using MEMD-PSO-SVR

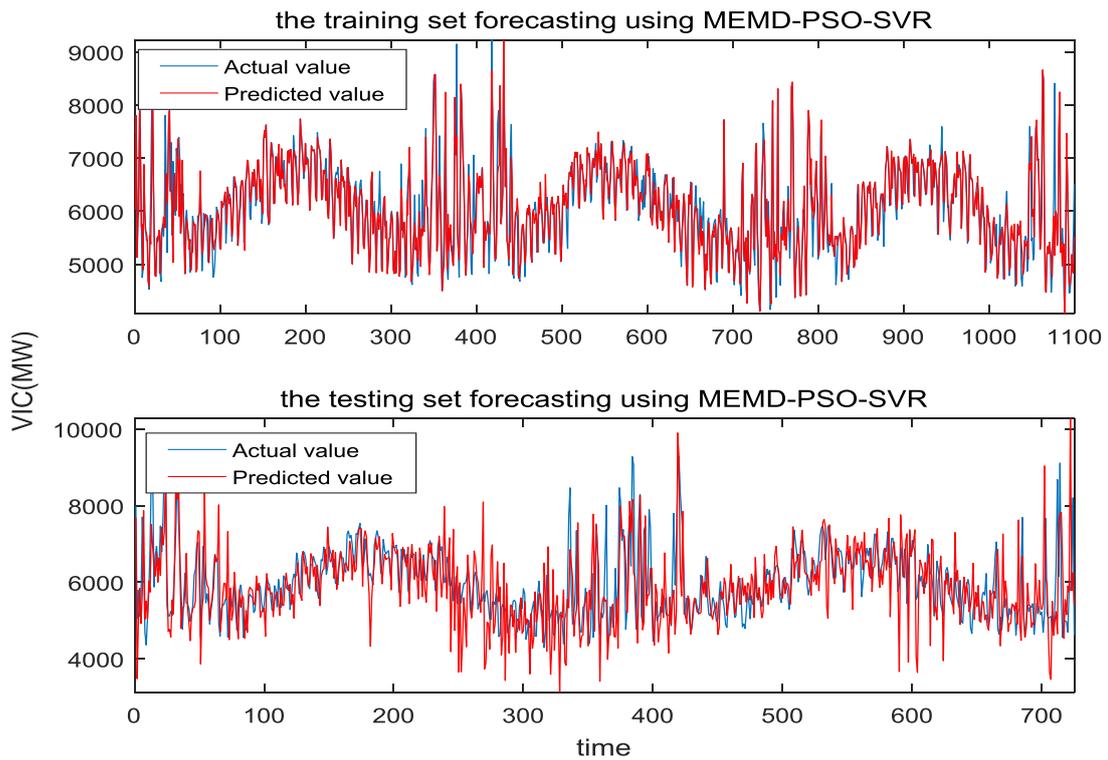

Fig. 10. The prediction of VIC using MEMD-PSO-SVR

## 5. Discussion

Our findings may serve as a foundation for developing a model using the multivariate time-frequency approach for predicting day-ahead peak load. The suggested model added two benefits: (1) it reduced the computational complexity required to predict day-ahead peak load accurately, and (2) it offered a robust peak load prediction model for the economic functioning of the power system.

Our first implementation procedures are as follows: to begin, meteorological factors would be taken into account. As mentioned in the introduction, predicting day-ahead peak loads is critical for the distribution and dispatch of power systems; yet, using the univariate historical load seems inadequate. It should be noted that accounting for all factors affecting peak load generates cumulative error and increases the computational complexity of hybrid models. As a result, temperature, a critical component in peak load forecasting's meteorological factors, has been included as an input variable in this study. Second, integrating several techniques into a hybrid model seemed to be a realistic choice to enhance the prediction accuracy. While single SVR and BPNN models can account for additional load affecting variables, prediction performance is still suboptimal. In light of this, we included the time-frequency analysis method into our study. In contrast to previous decomposition techniques, MEMD exploits the non-linearity and non-stationarity of multi-dimensional time series to concurrently extract inherent information obtained from the raw time series obtained through signal decomposition. According to Section 2.1, the MEMD technique employs multivariate data decomposition to efficiently minimize information loss during multivariate time series decomposition. As shown in Table 6, the time-frequency decomposition approach may substantially enhance peak load forecast accuracy. Additionally, to provide a more complete description of the MEMD method and its incorporation into our proposed model, we used the EEMD method without temperature data. Notably, since EEMD is incapable of handling multivariate time series data, we used univariate time series modeling with SVR (with and without optimization) to understand the difference between univariate and multivariate time series modeling. While the prediction accuracy of the EEMD-SVR model is superior to that of a single SVR model, even when temperature is included in the SVR model. Additionally, the MEMD-SVR approach with temperature variable produces more superior results in terms of all error metrics when compared to the EEMD-SVR model without temperature variable.

Additionally, the following strengths are our second implementation steps: to begin, a robust theoretical foundation and the capacity to generalize should be considered for modeling and forecasting. BPNN is an example of an ANN that is based on an empirical risk minimization approach, while SVR is an example of a structural risk minimization approach. As described in Section 2.2, theoretically SVR has more effective generalization ability than BPNN. While ANNs may be used to predict energy demand, they still have the disadvantages of leading to local optimum and over-fitting. As illustrated in Tables 6 and Table 7, BPNN-based approaches consistently perform worse than SVR-based models across all assessment criteria, despite the fact that the MAPE and RMSE of BPNN are better than those of SVR due to the adaptive learning of neural networks. Furthermore, as stated in Section 2.3, to ensure the forecasting accuracy of day-ahead peak load, PSO is used to carefully tune three hyper-parameters of SVR, namely the penalty parameter $C$, the width of insensitive loss function $\varepsilon$, and the RBF kernel parameter $\gamma$. To the forecasting accuracy of electricity peak load, PSO is employed to optimize three hyper-parameters of SVR. SVR is used to establish a model with different optimal parameters ($C, \varepsilon, \gamma$) for each multi-dimensional component decomposed by MEMD. As shown in Table 6 and Table 7, the proposed MEMD-PSO-SVR model attains better results in all evaluation criterions with SVR's parameters optimization using PSO than MEMD-SVR without optimizing the SVR's parameters.

Nonetheless, we would like to highlight the superiorities of the proposed MEMD-PSO-SVR model for day-ahead peak load forecasting. First, the MEMD technique can comprehensively capture the non-linear and non-stationary feature of multi-dimensional variables based on adaptively time-frequency analysis method. Secondly, SVR using RBF as the kernel function only has three parameters to be optimized by PSO, and could overcome the shortcoming of multi parameters in BPNN. Finally, as shown in Tables 6-7, the proposed MEMD-PSO-SVR model demonstrates higher performance in terms of DA, MAPE, RMSE, and $R^2$, and the DM test values and P-values between the proposed model and other six models further illustrate the proposed model's superiority.

## 6. Conclusions

To improve the forecasting accuracy of electrical load and decrease the computational cost of a forecasting model, a time-frequency analysis technique is extremely important in peak load forecasting. In this study, we have proposed a MEMD based hybrid method to settle the time-frequency problem. The proposed MEMD based hybrid method cooperates with a parameters' optimization method and a modeling technique. The MEMD algorithm is firstly used to simultaneously decompose the multivariate input variables, including the peak load, the valley, the mean load and temperature. To obtain the better performance, a PSO based parameters optimization is then employed to search for the optimal parameters of SVR. Then the models are trained to forecast each multivariate component extracted by MEMD. The forecast of each component is agglomerated into the final forecast value by another SVR model. The experimental results and statistical analysis demonstrated that the proposed hybrid modeling framework outperforms six other comparable methods.

Accordingly, the proposed hybrid model can become a suitable modeling framework in peak load forecasting. However, we considered only the historical load and the corresponding temperature as input variables of load forecasting. Some other lagged variables and more exogenous variables might be examined for improving the forecasting performance. Additionally, PSO and its variants are usually time-consuming to search for the optimal parameters of SVM. Therefore, some heuristic algorithms should be called upon for addressing the computational cost in the further investigation.

**Acknowledgement**

This work was supported by Humanities and Social Sciences Projects of Jiangxi under project no.GL19115, Jiangxi Principal Academic and Technical Leaders Program under project no.20194BCJ22015, Science and Technology Research Projects of Jiangxi under project no.GJJ202905, and Natural Science Foundation of China under project Nos. 71571080 and 71871101.